\documentclass[11pt]{article}
\usepackage{coling2018}
\usepackage{times}
\usepackage{url}
\usepackage{bm}
\usepackage{multirow}
\usepackage{latexsym}
\usepackage{array}
\usepackage{comment}
\usepackage{color}
\newcolumntype{P}[1]{>{\centering\arraybackslash}p{#1}}
\usepackage{graphicx}
\usepackage{amssymb}
\usepackage{amsmath}
\usepackage{subcaption}
\usepackage{footnote}
\usepackage{floatrow}
\makesavenoteenv{tabular}
\newfloatcommand{capbtabbox}{table}[][\FBwidth]
\usepackage[letterpaper, margin=1in]{geometry}
\usepackage{booktabs}
\usepackage[toc,page]{appendix}

\title{Neural Network Models for Paraphrase Identification, Semantic Textual Similarity, Natural Language Inference, and Question Answering}

\author{}
\author{Wuwei Lan and Wei Xu  \\
Department of Computer Science and Engineering \\
  Ohio State University \\
  {\tt \{lan.105, xu.1265\}@osu.edu} \\}

\date{}

\begin{document}
\maketitle
\begin{abstract}
In this paper, we analyze several neural network designs (and their variations) for sentence pair modeling and compare their performance extensively across eight datasets, including paraphrase identification, semantic textual similarity, natural language inference, and question answering tasks. Although most of these models have claimed state-of-the-art performance, the original papers often reported on only one or two selected datasets. We provide a systematic study and show that (i) encoding contextual information by LSTM and inter-sentence interactions are critical, (ii) Tree-LSTM does not help as much as previously claimed but surprisingly improves performance on Twitter datasets, (iii) the Enhanced Sequential Inference Model \cite{Chen-Qian:2017:ACL} is the best so far for larger datasets, while the Pairwise Word Interaction Model \cite{he-lin:2016:N16-1} achieves the best performance when less data is available. We release our implementations as an open-source toolkit.
\end{abstract}

\section{Introduction}
\label{intro}

\blfootnote{
    %
    \hspace{-0.65cm}  
    This work is licensed under a Creative Commons Attribution 4.0 International License. License details: \url{http://creativecommons.org/licenses/by/4.0/}.
}

Sentence pair modeling is a fundamental technique underlying many NLP tasks, including the following:
 \begin{itemize}
   \item Semantic Textual Similarity (STS), which measures the degree of equivalence in the underlying semantics of paired snippets of text \cite{agirre-EtAl:2016:SemEval2}.
   \vspace{-.1in}
   \item Paraphrase Identification (PI), which identifies whether two sentences express the same meaning \cite{dolan2005automatically,Xu-EtAl-2014:TACL,xu2015semeval}. 
   \vspace{-.1in}
   \item Natural Language Inference (NLI), also known as recognizing textual entailment (RTE), which concerns whether a hypothesis can be inferred from a premise, requiring understanding of the semantic similarity between the hypothesis and the premise  \cite{Dagan:2005:PRT:2100045.2100054,snli:emnlp2015}.
   \vspace{-.1in}
   \item Question Answering (QA), which can be approximated as ranking candidate answer sentences or phrases based on their similarity to the original question \cite{yang-yih-meek:2015:EMNLP}.
   \vspace{-.1in}
   \item Machine Comprehension (MC), which requires sentence matching between a passage and a question, pointing out the text region that contains the answer. \cite{rajpurkar-EtAl:2016:EMNLP2016}.
\end{itemize}
Traditionally, researchers had to develop different methods specific for each task. Now neural networks can perform all the above tasks with the same architecture by training end to end. Various neural models \cite{he-lin:2016:N16-1,Chen-Qian:2017:ACL,parikh-EtAl:2016:EMNLP2016,wieting2016towards,tomar-EtAl:2017:SCLeM,wang2017bilateral,shen-yang-deng:2017:EMNLP2017,TACL831} have declared state-of-the-art results for sentence pair modeling tasks; however, they were carefully designed and evaluated on selected (often one or two) datasets that can demonstrate the superiority of the model. The research questions are as follows: Do they perform well on other tasks and datasets? How much performance gain is due to certain system design choices and hyperparameter optimizations?  

To answer these questions and better understand different network designs, we systematically analyze and compare the state-of-the-art neural models across multiple tasks and multiple domains. Namely, we implement five models and their variations on the same PyTorch platform: InferSent model \cite{conneau-EtAl:2017:EMNLP2017}, Shortcut-stacked Sentence Encoder Model \cite{nie-bansal:2017:RepEval}, Pairwise Word Interaction Model \cite{he-lin:2016:N16-1}, Decomposable Attention Model \cite{parikh-EtAl:2016:EMNLP2016}, and Enhanced Sequential Inference Model \cite{Chen-Qian:2017:ACL}. They are representative of the two most common approaches: \textbf{sentence encoding models} that learn vector representations of individual sentences and then calculate the semantic relationship between sentences based on vector distance and \textbf{sentence pair interaction models} that use some sorts of word alignment mechanisms (e.g., attention) then aggregate inter-sentence interactions. We focus on identifying important network designs and present a series of findings with quantitative measurements and in-depth analyses, including (i) incorporating inter-sentence interactions is critical; (ii) Tree-LSTM does not help as much as previously claimed but surprisingly improves performance on Twitter data; (iii) Enhanced Sequential Inference Model has the most consistent high performance for larger datasets, while Pairwise Word Interaction Model performs better on smaller datasets and Shortcut-Stacked Sentence Encoder Model is the best performaning model on the Quora corpus. We release our implementations as a toolkit to the research community.\footnote{The code is available on the authors' homepages and GitHub: \url{https://github.com/lanwuwei/SPM_toolkit}}

\section{General Framework for Sentence Pair Modeling}
\label{general_framework}

Various neural networks have been proposed for sentence pair modeling, all of which fall into two types of approaches. The sentence encoding approach encodes each sentence into a fixed-length vector and then computes sentence similarity directly. The model of this type has advantages in the simplicity of the network design and generalization to other NLP tasks. The sentence pair interaction approach takes word alignment and interactions between the sentence pair into account and often show better performance when trained on in-domain data. Here we outline the two types of neural networks under the same general framework:
 
\begin{itemize}
    \item \textbf{The Input Embedding Layer} takes vector representations of words as input, where pretrained word embeddings are most commonly used, e.g. GloVe \cite{pennington2014glove} or Word2vec \cite{mikolov2013distributed}. Some work used embeddings specially trained on phrase or sentence pairs that are paraphrases \cite{wieting-gimpel:2017:Long,tomar-EtAl:2017:SCLeM}; some used subword embeddings, which showed improvement on social media data \cite{lan2018subword}. 
    
    \vspace{-.1in}
    \item \textbf{The Context Encoding Layer} incorporates word context and sequence order into modeling for better vector representation. This layer often uses CNN \cite{he-gimpel-lin:2015:EMNLP}, LSTM \cite{Chen-Qian:2017:ACL}, recursive neural network \cite{socher2011parsing}, or highway network \cite{gong2017natural}. The sentence encoding type of model will stop at this step, and directly use the encoded vectors to compute the semantic similarity through vector distances and/or the output classification layer.
    
    \vspace{-.1in}
    \item \textbf{The Interaction and Attention Layer} calculates word pair (or n-gram pair) interactions using the outputs of the encoding layer. This is the key component for the interaction-aggregation type of model. In the PWIM model \cite{he-lin:2016:N16-1}, the interactions are calculated by cosine similarity, Euclidean distance, and the dot product of the vectors. Various models put different weights on different interactions, primarily simulating the word alignment between two sentences. The alignment information is useful for sentence pair modeling because the semantic relation between two sentences depends largely on the relations of aligned chunks as shown in the SemEval-2016 task of interpretable semantic textual similarity \cite{agirre-EtAl:2016:SemEval2}.  
    
    \vspace{-.1in}
    \item \textbf{The Output Classification Layer} adapts CNN or MLP to extract semantic-level features on the attentive alignment and applies softmax function to predict probability for each class.
\end{itemize}

\section{Representative Models for Sentence Pair Modeling}
\label{representation_models}
Table \ref{all_model_comparison} gives a summary of typical models for sentence pair modeling in recent years. In particular, we investigate five models in depth: two are representative of the sentence encoding type of model, and three are representative of the interaction-aggregation type of model. These models have reported state-or-the-art results with varied architecture design (this section) and implementation details (Section \ref{sec:implementation}). 

\begin{table}[ht!]
\footnotesize
\centering
\begin{tabular}{>{\centering\arraybackslash}m{4cm}>{\centering\arraybackslash}m{3.5cm}>{\centering\arraybackslash}m{4cm}>{\centering\arraybackslash}m{3cm}}
\hline
\hline
\multirow{2}{*}{\textbf{Models}} & \textbf{   Sentence} & \textbf{Interaction and} & \textbf{Aggregation and} \\ 
& \textbf{Encoder} & \textbf{Attention} & \textbf{Classification} \\
\hline 
\\ [-1.5ex]

\cite{shen2017disan} & Directional self-attention network & - & MLP \\ [1.5ex]

\cite{choi2017unsupervised} & Gumbel Tree-LSTM & - & MLP \\ [1.5ex]

\cite{wieting-gimpel:2017:Long} & Gated recurrent average network & - & MLP \\ [1.5ex]

\textbf{SSE} \cite{nie-bansal:2017:RepEval} & Shortcut-stacked BiLSTM & - & MLP \\ [1.5ex]

\hline
\\ [-1.5ex]

\cite{he-gimpel-lin:2015:EMNLP} & CNN & multi-perspective matching & pooling + MLP \\ [1.5ex]

\cite{rocktaschel2015reasoning} & LSTM & word-by-word neural attention & MLP \\ [1.5ex]

\cite{liu2016modelling} & LSTM & coupled LSTMs & dynamic pooling + MLP \\ [1.5ex]

\cite{TACL831} & CNN & attention matrix & logistic regression\\ [1.5ex]

\textbf{DecAtt} \cite{parikh-EtAl:2016:EMNLP2016} & - & dot product + soft alignment & summation + MLP \\ [1.5ex]

\textbf{PWIM} \cite{he-lin:2016:N16-1} & BiLSTM & cosine, Euclidean, dot product + hard alignment  & CNN + MLP \\ [1.5ex]

\cite{wang2016compare} & LSTM encodes both context and attention & word-by-word neural attention & MLP \\ [1.5ex]

\textbf{ESIM} \cite{Chen-Qian:2017:ACL} & BiLSTM (Tree-LSTM) before and after attention & dot product + soft alignment & average and max pooling + MLP\\ [1.5ex]

\cite{wang2017bilateral} & BiLSTM & multi-perspective matching & BiLSTM + MLP \\ [1.5ex]

\cite{shen-yang-deng:2017:EMNLP2017} & BiLSTM + intra-attention & soft alignment + orthogonal decomposition & MLP \\ [1.5ex]

\cite{ghaeini2018dr} & dependent reading BiLSTM & dot product + soft alignment & average and max pooling+MLP\\ [1.5ex]

\hline
\end{tabular}
\caption{Summary of representative neural models for sentence pair modeling. The upper half contains sentence encoding models, and the lower half contains sentence pair interaction models.} 
\label{all_model_comparison}
\end{table}

\begin{figure}[bp!]
\vspace{-.4in}
\begin{subfigure}{0.29\textwidth}
\includegraphics[width=\linewidth]{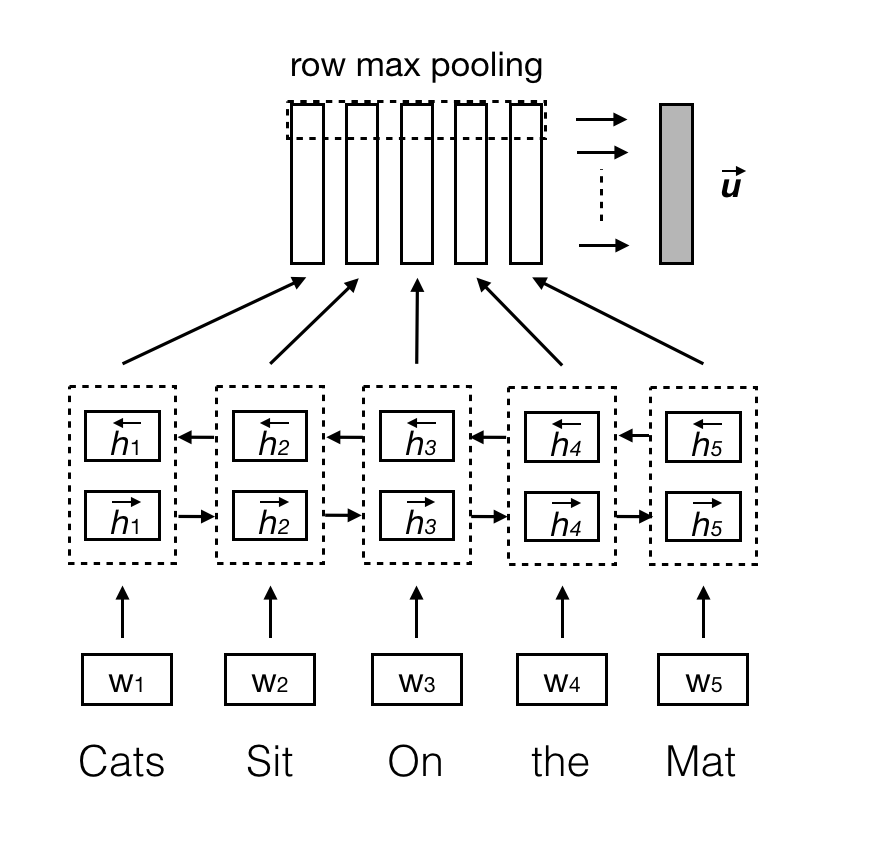} 
\caption{InferSent}
\end{subfigure}
\begin{subfigure}{0.29\textwidth}
\includegraphics[width=\linewidth]{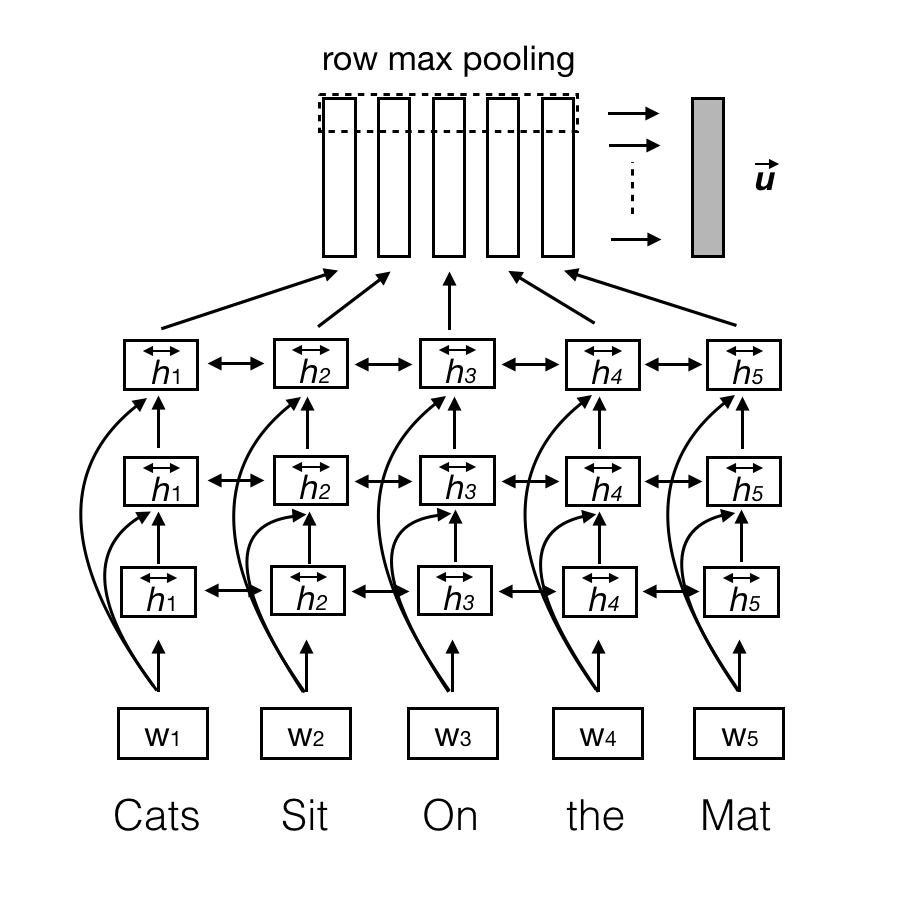}
\caption{SSE}
\end{subfigure}
\begin{subfigure}{0.40\textwidth}
\includegraphics[width=\linewidth]{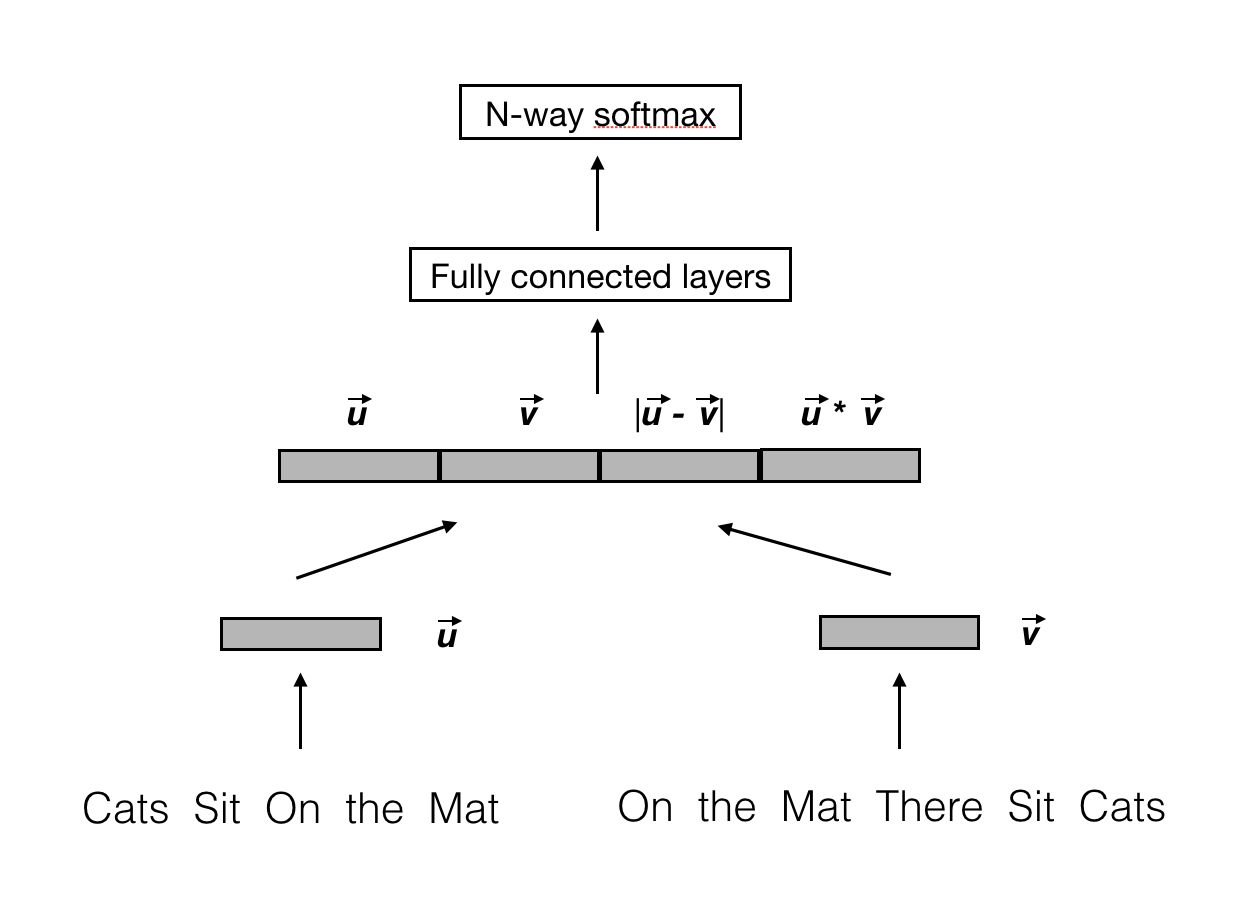}
\caption{Classification Layer}
\end{subfigure}

\caption{\textbf{Sentence encoding models} focus on learning vector representations of individual sentences and then calculate the semantic relationship between sentences based on vector distance.}
\label{fig:model_figure}
\end{figure}

\subsection{The Bi-LSTM Max-pooling Network (InferSent)}
\label{sec:InferSent}

We choose the simple Bi-LSTM max-pooling network from InferSent \cite{conneau-EtAl:2017:EMNLP2017}:
\begin{align}
    &\overleftrightarrow{\bm{h}}_{i} = BiLSTM(\bm{x}_{i}, \overleftrightarrow{\bm{h}}_{i-1}) \\
    &\bm{v}=max(\overleftrightarrow{\bm{h}}_{1}, \overleftrightarrow{\bm{h}}_{2}, ..., \overleftrightarrow{\bm{h}}_{n})
\end{align}
\noindent where \(\overleftrightarrow{\bm{h}}_{i}\) represents the concatenation of hidden states in both directons. It has shown better transfer learning capabilities than several other sentence embedding models, including SkipThought \cite{NIPS2015_5950} and FastSent \cite{hill-cho-korhonen:2016:N16-1}, when trained on the natural language inference datasets. 

\subsection{The Shortcut-Stacked Sentence Encoder Model (SSE)}
\label{sec:SSE_model}
The Shortcut-Stacked Sentence Encoder model \cite{nie-bansal:2017:RepEval} is a sentence-based embedding model, which enhances multi-layer Bi-LSTM with skip connection to avoid training error accumulation, and calculates each layer as follows:
\begin{align}
    &\overleftrightarrow{\bm{h}}_{i}^{k} = BiLSTM(\bm{x}_{i}^{k}, \overleftrightarrow{\bm{h}}_{i-1}^{k}) \\
    &\bm{x}_{i}^{1}=\bm{w}_{i} \quad (k=1), \qquad 
    \bm{x}_{i}^{k}=[\bm{w}_{i}, \overleftrightarrow{\bm{h}}_{i}^{k-1}, \overleftrightarrow{\bm{h}}_{i}^{k-2}, ..., \overleftrightarrow{\bm{h}}_{i}^{1}] \quad (k>1) \\
    &\bm{v}=max(\overleftrightarrow{\bm{h}}_{1}^{m}, \overleftrightarrow{\bm{h}}_{2}^{m}, ..., \overleftrightarrow{\bm{h}}_{n}^{m})
\end{align}
\noindent where \(\bm{x}_{i}^{k}\) is the input of the $k$th Bi-LSTM layer at time step $i$, which is the combination of outputs from all previous layers, \(\overleftrightarrow{\bm{h}}_{i}^{k}\) represents the hidden state of the $k$th Bi-LSTM layer in both directions. The final sentence embedding $\bm{v}$ is the row-based max pooling over the output of the last Bi-LSTM layer, where $n$ denotes the number of words within a sentence and $m$ is the number of Bi-LSTM layers ($m = 3$  in SSE).

\subsection{The Pairwise Word Interaction Model (PWIM)}
\label{sec:PWIM_model}
In the Pairwise Word Interaction model \cite{he-lin:2016:N16-1}, each word vector \(\bm{w}_{i}\) is encoded with context through forward and backward LSTMs: $\overrightarrow{\bm{h}}_{i} = LSTM^{f}(\bm{w}_{i}, \overrightarrow{\bm{h}}_{i-1})$ and $\overleftarrow{\bm{h}}_{i} = LSTM^{b}(\bm{w}_{i}, \overleftarrow{\bm{h}}_{i+1})$. For every word pair $(\bm{w}^a_{i}, \bm{w}^b_{j})$ across sentences, the model directly calculates word pair interactions using cosine similarity, Euclidean distance, and dot product over the outputs of the encoding layer:
\begin{align}
    D(\overrightarrow{\bm{h}}_{i}, \overrightarrow{\bm{h}}_{j}) & = [cos(\overrightarrow{\bm{h}}_{i}, \overrightarrow{\bm{h}}_{j}),  \|\overrightarrow{\bm{h}}_{i} - \overrightarrow{\bm{h}}_{j}\|, \overrightarrow{\bm{h}}_{i} \cdot \overrightarrow{\bm{h}}_{j}] 
\end{align}
\noindent The above equation not only applies to forward hidden state \(\overrightarrow{\bm{h}}_{i}\) and backward hidden state \(\overleftarrow{\bm{h}}_{i}\), but also to the concatenation \(\overleftrightarrow{\bm{h}}_{i}= [\overrightarrow{\bm{h}}_{i}, \overleftarrow{\bm{h}}_{i}]\) and summation \(\bm{h}^{+}_{i}= \overrightarrow{\bm{h}}_{i} + \overleftarrow{\bm{h}}_{i}\), resulting in a tensor \(\mathbf{D}^{13 \times |sent1| \times |sent2|}\) after padding one extra bias term. A ``hard'' attention is applied to the interaction tensor to build word alignment: selecting the most related word pairs and increasing the corresponding weights by 10 times. Then a 19-layer deep CNN is applied to aggregate the word interaction features for final classification.

\subsection{The Decomposable Attention Model (DecAtt)}
\label{sec:DecAtt_model}
The Decomposable Attention model \cite{parikh-EtAl:2016:EMNLP2016} is one of the earliest models to introduce attention-based alignment for sentence pair modeling, and it achieved state-of-the-art results on the SNLI dataset with about an order of magnitude fewer parameters than other models (see more in Table \ref{tab:training_timel}) without relying on word order information. It computes the word pair interaction between \(\bm{w}_{i}^{a}\) and \(\bm{w}_{j}^{b}\) (from input sentences $s_a$ and $s_b$, each with $m$ and $n$ words, respectively) as \({e}_{ij} = {F(\bm{w}_{i}^{a})}^{T}  F(\bm{w}_{j}^{b})\), where \(F\) is a feedforward network; then alignment is determined as follows:
\begin{align}
    &\bm{\beta}_{i} = \sum_{j=1}^{n} \frac{exp({e}_{ij})}{\sum_{k=1}^{n} exp({e}_{ik})} \bm{w}_{j}^{b} \quad
    &\bm{\alpha}_{j} = \sum_{i=1}^{m} \frac{exp({e}_{ij})}{\sum_{k=1}^{m} exp({e}_{kj})} \bm{w}_{i}^{a} \label{eq:soft_alignment}
\end{align}

\noindent where \(\bm{\beta}_{i}\) is the soft alignment between \(\bm{w}_{i}^{a}\) and subphrases  \(\bm{w}_{j}^{b}\) in sentence $s_b$, and vice versa for \(\bm{\alpha}_{j}\). The aligned phrases are fed into another feedforward network \(G\): \(\bm{v}_{i}^{a} = G([\bm{w}_{i}^{a}; \bm{\beta}_{i}]) \label{eq:alignment_collection}\) and \(\bm{v}_{j}^{b} = G([\bm{w}_{j}^{b}; \bm{\alpha}_{j}]) \label{eq:alignment_collection2}\) to generate sets \(\{\bm{v}_{i}^{a}\}\) and \(\{\bm{v}_{j}^{b}\}\), which are aggregated by summation and then concatenated together for classification.

\subsection{The Enhanced Sequential Inference Model (ESIM)}
\label{sec: ESIM_model}
The Enhanced Sequential Inference Model \cite{Chen-Qian:2017:ACL} is closely related to the DecAtt model, but it differs in a few aspects. First, Chen et al. \shortcite{Chen-Qian:2017:ACL} demonstrated that using Bi-LSTM to encode sequential contexts is important for performance improvement. They used the concatenation \(\overline{\bm{w}}_{i} = \overleftrightarrow{\bm{h}}_{i} = [\overrightarrow{\bm{h}}_{i}, \overleftarrow{\bm{h}}_{i}]\) of both directions as in the PWIM model. The word alignment \(\bm{\beta}_{i}\) and \(\bm{\alpha}_{j}\) between \(\overline{\bm{w}}^{a}\) and \(\overline{\bm{w}}^{b}\) are calculated the same way as in DecAtt. Second, they showed the competitive performance of recursive architecture with constituency parsing, which complements with sequential LSTM. The feedforward function \(G\) in DecAtt is replaced with Tree-LSTM:
\begin{align}
    &\bm{v}_{i}^{a} = TreeLSTM([\overline{\bm{w}}_{i}^{a}; \bm{\beta}_{i}; \overline{\bm{w}}_{i}^{a}-\bm{\beta}_{i}; \overline{\bm{w}}_{i}^{a} \odot \bm{\beta}_{i}])\\
    &\bm{v}_{j}^{b} = TreeLSTM([\overline{\bm{w}}_{j}^{b}; \bm{\alpha}_{j}; \overline{\bm{w}}_{j}^{b}-\bm{\alpha}_{j}; \overline{\bm{w}}_{j}^{b} \odot \bm{\alpha}_{j}])
\end{align}
Third, instead of using summation in aggregation, ESIM adapts the average and max pooling and concatenation \(\bm{v}= [\bm{v}_{ave}^{a}; \bm{v}_{max}^{a}; \bm{v}_{ave}^{b}; \bm{v}_{max}^{b}]\) before passing through multi-layer perceptron (MLP) for classification: 
\begin{align}
    & \bm{v}_{ave}^{a} = \sum_{i=1}^{m} \frac{\bm{v}_{i}^{a}}{m},  \qquad \bm{v}_{max}^{a} = \max_{i=1}^{m}\bm{v}_{i}^{a} , \qquad  \bm{v}_{ave}^{b} = \sum_{j=1}^{n} \frac{\bm{v}_{j}^{b}}{n}, \qquad \bm{v}_{max}^{b} = \max_{j=1}^{n}\bm{v}_{j}^{b} 
\end{align}

\begin{figure}[ht!]
\vspace{-.2in}
\begin{subfigure}{0.48\textwidth}
\includegraphics[width=\linewidth]{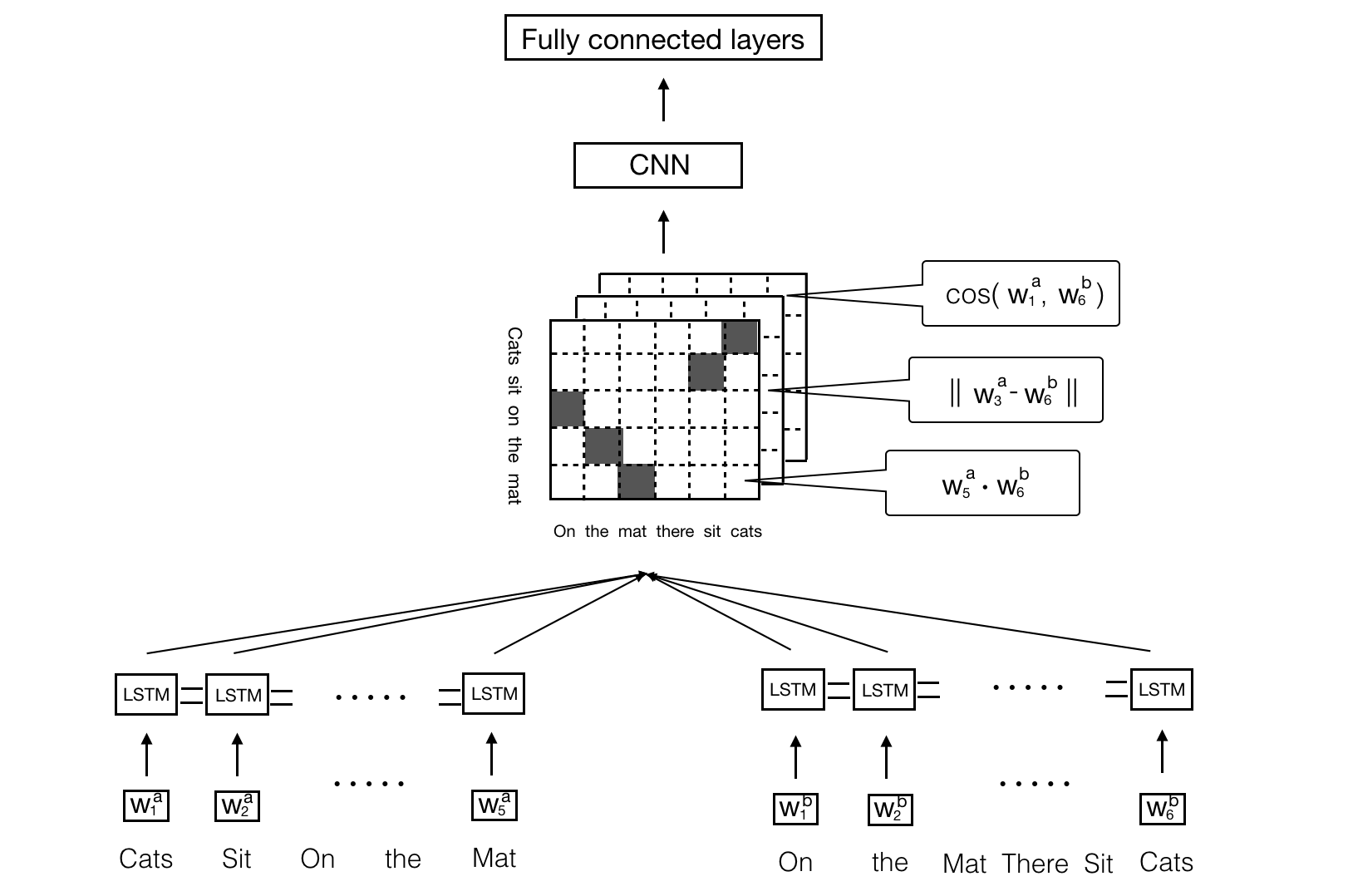}
\caption{PWIM}
\end{subfigure}
\begin{subfigure}{0.48\textwidth}
\includegraphics[width=\linewidth]{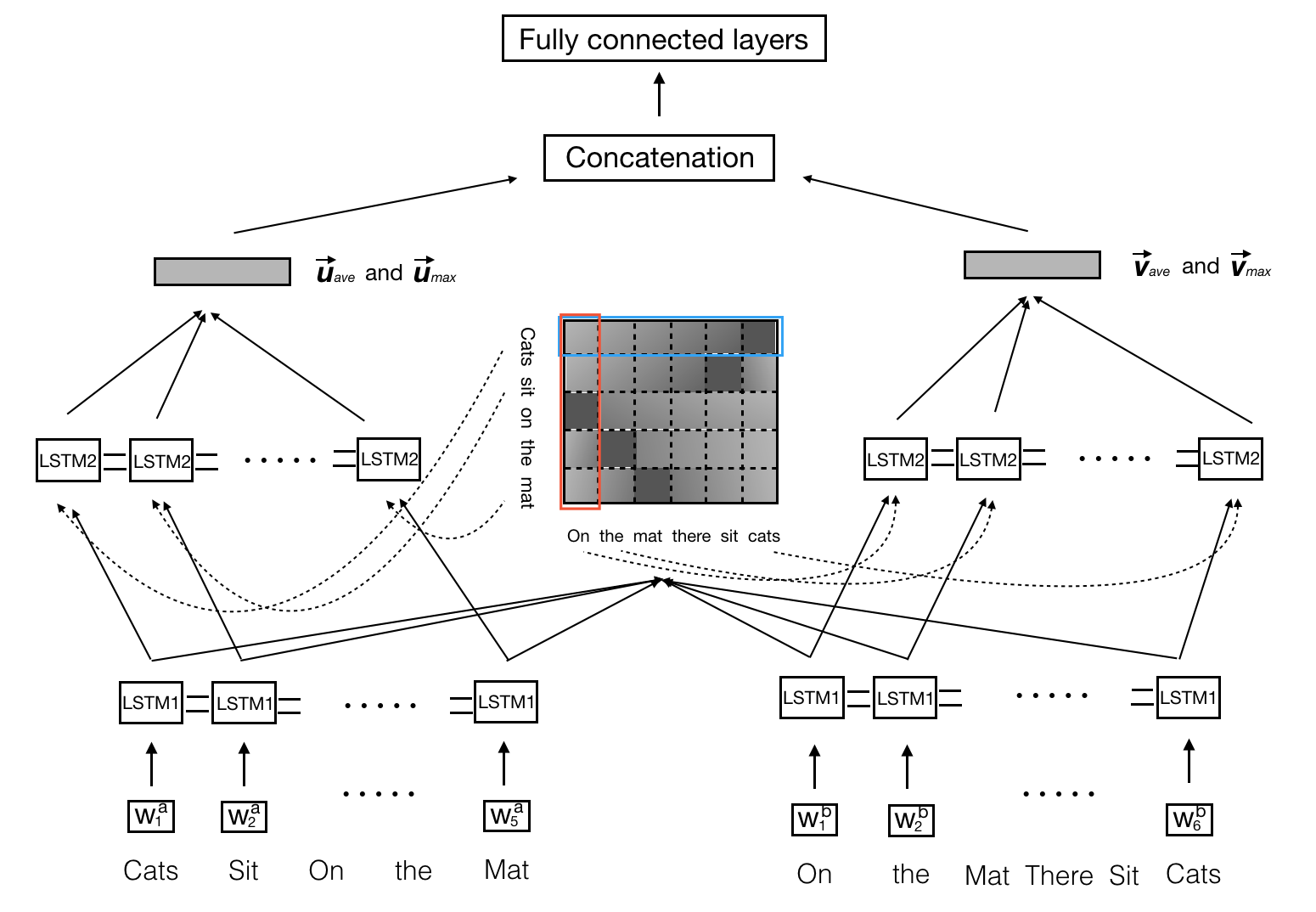}
\caption{ESIM$_{seq}$ (DecAtt is similar and simpler.)}
\end{subfigure}
\caption{\textbf{Sentence pair interaction models} use different word alignment mechanisms before aggregation.}
\end{figure}

\section{Experiments and Analysis}

\subsection{Datasets}
We conducted sentence pair modeling experiments on eight popular datasets: two NLI datasets, three PI datasets, one STS dataset and two QA datasets. Table \ref{all_data_comparison} gives a comparison of these datasets:
\begin{table*}[ht!]
\small
\centering
\begin{tabular}{P{1.5cm}P{0.4cm}P{1cm}P{7.8cm}P{1.7cm}}
 \hline
 \hline
 \\
 \textbf{Dataset} & \multicolumn{2}{c}{\textbf{Size}} & \multicolumn{2}{c}{\textbf{Example and Label}}\\ [1.2ex]
 \hline
 \multirow{3}{*}{SNLI} & train & 550,152 & \multirow{2}{*}{\textit{$s_a$: Two men on bicycles competing in a race.}} & \textbf{entailment} \\
 & dev & 10,000 & \multirow{2}{*}{\textit{$s_b$: Men are riding bicycles on the street.}} & neutral \\
 & test & 10,000 &  & contradict \\
 \hline
 \multirow{3}{*}{Multi-NLI} & train & 392,703 & \multirow{2}{*}{\textit{$s_a$: The Old One always comforted Ca'daan, except today.}} & entailment\\
 & dev & 20,000 & \multirow{2}{*}{\textit{$s_b$: Ca'daan knew the Old One very well.}} & \textbf{neutral} \\
 & test & 20,000 &  & contradict \\
 \hline
 \multirow{3}{*}{Quora} & train & 384,348 &  \multirow{2}{*}{\textit{$s_a$: What should I do to avoid sleeping in class?}} & \multirow{2}{*}{\textbf{paraphrase}}\\
 & dev & 10,000 & \multirow{2}{*}{\textit{$s_b$: How do I not sleep in a boring class?}} &  \multirow{2}{*}{non-paraphrase}  \\
 & test & 10,000 &  \\
 \hline
 \multirow{3}{*}{Twitter-URL} & train & 42,200 & \multirow{2}{*}{\textit{$s_a$: Letter warned Wells Fargo of ``widespread'' fraud in 2007.}} & \multirow{2}{*}{\textbf{paraphrase}}\\
 & dev & - & \multirow{2}{*}{\textit{$s_b$: Letters suggest Wells Fargo scandal started earlier.}} & \multirow{2}{*}{non-paraphrase}\\
 & test & 9,324 & \\
 \hline
 \multirow{3}{*}{PIT-2015} & train & 11,530 & 
 \multirow{2}{*}{\textit{$s_a$: Ezekiel Ansah w the 3D shades Popped out lens }} & \multirow{2}{*}{\textbf{paraphrase}} \\
 & dev & 4,142 & \multirow{2}{*}{\textit{$s_b$: Ezekiel Ansah was wearing lens less 3D glasses}} & \multirow{2}{*}{non-paraphrase} \\
 & test & 838 & \\
 \hline
 \multirow{3}{*}{STS-2014}& train & 7,592 & \multirow{2}{*}{\textit{$s_a$: Then perhaps we could have avoided a catastrophe.}} & \multirow{2}{*}{score \([0, 5]\)}\\
 & dev & - & \multirow{2}{*}{\textit{$s_b$: Then we might have been able to avoid a disaster.}} & \multirow{2}{*}{\(\bm{4.6}\)} \\
 & test & 3,750 &   \\
 \hline
 \multirow{3}{*}{WikiQA} & train & 8,672 & \multirow{2}{*}{\textit{$s_a$: How much is 1 tablespoon of water?}}  & \multirow{2}{*}{\textbf{true}}\\
 & dev & 1,130 & \multirow{2}{*}{\textit{$s_b$: In Australia one tablespoon (measurement unit) is 20 mL.}} & \multirow{2}{*}{false}\\
 & test & 2,351 &   \\
 \hline
 \multirow{3}{*}{TrecQA} & train & 53,417 & \multirow{2}{*}{\textit{$s_a$: Who was Lincoln's Secretary of State?}} & \multirow{2}{*}{true} \\
 & dev & 1,148 & \multirow{2}{*}{\textit{$s_b$: William Seward}} & \multirow{2}{*}{\textbf{false}} \\
 & test & 1,517 &  \\
 \hline

\end{tabular}
\caption{Basic statistics and examples of different datasets for sentence pair modeling tasks.} 
\label{all_data_comparison}
\end{table*}

\vspace{-.1in}
\begin{itemize}
    \item \textbf{SNLI} \cite{snli:emnlp2015} contains 570k hypotheses written by crowdsourcing workers given the premises. It focuses on three semantic relations: the premise entails the hypothesis (entailment), they contradict each other (contradiction), or they are unrelated (neutral). 
    \vspace{-.1in}
    \item \textbf{Multi-NLI} \cite{williams2017broad} extends the SNLI corpus to multiple genres of written and spoken texts with 433k sentence pairs. 
    \vspace{-.1in}
    \item \textbf{Quora} \cite{Quora} contains 400k question pairs collected from the Quora website. This dataset has balanced positive and negative labels indicating whether the questions are duplicated or not. 
    \vspace{-.1in}
    \item \textbf{Twitter-URL} \cite{lan2017continuously} includes 50k sentence pairs collected from tweets that share the same URL of news articles. This dataset contains both formal and informal language. 
    \vspace{-.1in}
    \item \textbf{PIT-2015} \cite{xu2015semeval} comes from SemEval-2015 and was collected from tweets under the same trending topic. It contains naturally occurred (i.e. written by independent Twitter users spontaneously) paraphrases and non-paraphrases with varied topics and language styles. 
    
    \vspace{-.1in}
    \item \textbf{STS-2014} \cite{agirre-EtAl:2014:SemEval} is from SemEval-2014, constructed from image descriptions, news headlines, tweet news, discussion forums, and OntoNotes \cite{hovy-EtAl:2006:HLT-NAACL06-Short}. 
    \vspace{-.1in}
    \item \textbf{WikiQA} \cite{yang-yih-meek:2015:EMNLP} is an open-domain question-answering dataset. Following He and Lin \shortcite{he-lin:2016:N16-1}, questions without correct candidate answer sentences are excluded, and answer sentences are truncated to 40 tokens, resulting in 12k question-answer pairs for our experiments. 
    \vspace{-.1in}
    \item \textbf{TrecQA} \cite{wang2007jeopardy} is an answer selection task of 56k question-answer pairs and created in Text Retrieval Conferences (TREC). For both WikiQA and TrecQA datasets, the best answer is selected according to the semantic relatedness with the question. 
\end{itemize}

\subsection{Implementation Details}
\label{sec:implementation} 
We implement all the models with the same PyTorch framework.\footnote{InferSent and SSE have open-source PyTorch implementations by the original authors, for which we reused part of the code.}\footnote{Our code is available at: \url{https://github.com/lanwuwei/SPM_toolkit}} Below, we summarize the implementation details that are key for reproducing results for each model:
\begin{itemize}
    \vspace{-.1in}
    \item \textbf{SSE:} This model can converge very fast, for example, 2 or 3 epochs for the SNLI dataset. We control the convergence speed by updating the learning rate for each epoch: specifically, \(lr=\frac{1}{2^{\frac{epoch\_i}{2}}}*{init\_lr}\), where \(init\_lr\) is the initial learning rate and \(epoch\_i\) is the index of current epoch. 
    \vspace{-.1in}
    \item \textbf{DecAtt:}
    It is important to use gradient clipping for this model: for each gradient update, we check the L2 norm of all the gradient values, if it is greater than a threshold \(b\), we scale the gradient by a factor \(\alpha = b/L2\_norm\). Another useful procedure is to assemble batches of sentences with similar length. 
    \vspace{-.1in}
    \item \textbf{ESIM:} Similar but different from DecAtt, ESIM batches sentences with varied length and uses masks to filter out padding information. In order to batch the parse trees within Tree-LSTM recursion, we follow Bowman et al.'s \shortcite{bowman-EtAl:2016:P16-1} procedure that converts tree structures into the linear sequential structure of a shift reduce parser. Two additional masks are used for producing left and right children of a tree node.
    \vspace{-.1in}
    \item \textbf{PWIM:} The cosine and Euclidean distances used in the word interaction layer have smaller values for similar vectors while dot products have larger values. The performance increases if we add a negative sign to make all the vector similarity measurements behave consistently.
\end{itemize}

\subsection{Analysis}

\subsubsection{Re-implementation Results vs. Previously Reported Results}
Table \ref{model_reported_results} and \ref{model_replicated_results} show the results reported in the original papers and the replicated results with our implementation. We use accuracy, F1 score, Pearson's $r$, Mean Average Precision (MAP), and Mean Reciprocal Rank (MRR) for evaluation on different datasets following the literature. Our reproduced results are slightly lower than the original results by 0.5 $\sim$ 1.5 points on accuracy. We suspect the following potential reasons: (i) less extensive  hyperparameter tuning for each individual dataset; (ii) only one run with random seeding to report results; and (iii) use of different neural network toolkits: for example, the original ESIM model was implemented with Theano, and PWIM model was in Torch. 

\subsubsection{Effects of Model Components}
Herein, we examine the main components that account for performance in sentence pair modeling.\\

\vspace{-.1in}
\noindent\textbf{How important is LSTM encoded context information for sentence pair modeling?}\\
Regarding DecAtt, Parikh et al. \shortcite{parikh-EtAl:2016:EMNLP2016} mentioned that ``intra-sentence attention is optional"; they can achieve competitive results without considering context information. However, not surprisingly, our experiments consistently show that encoding sequential context information with LSTM is critical. Compared to DecAtt, ESIM shows better performance on every dataset (see Table \ref{model_replicated_results} and Figure \ref{fig:Training_curve}). The main difference between ESIM and DecAtt that contributes to performance improvement, we found, is the use of Bi-LSTM and Tree-LSTM for sentence encoding, rather than the different choices of aggregation functions. \\

\vspace{-.1in}
\noindent\textbf{Why does Tree-LSTM help with Twitter data?}\\
Chen et al. \shortcite{Chen-Qian:2017:ACL} offered a simple combination (ESIM$_{seq+tree}$) by averaging the prediction probabilities of two ESIM variants that use sequential Bi-LSTM and Tree-LSTM respectively, and suggested ``parsing information complements very well with ESIM and further improves the performance". However, we found that adding Tree-LSTM only helps slightly or not at all for most datasets, but it helps noticably with the two Twitter paraphrase datasets. We hypothesize the reason is that these two datasets come from real-world tweets which often contain extraneous text fragments, in contrast to SNLI and other datasets that have sentences written by crowdsourcing workers. For example, the segment ``\textit{ever wondered ,}'' in the sentence pair \textit{ever wondered , why your recorded \#voice sounds weird to you?} and \textit{why do our recorded voices sound so weird to us?} introduces a disruptive context into the Bi-LSTM encoder, while Tree-LSTM can put it in a less important position after constituency parsing.\\


\bgroup
\def\arraystretch{1.8}
\begin{table*}[ht!]
\small
\centering
\begin{tabular}{P{1.6cm}P{1cm}P{1.6cm}P{1cm}P{1.8cm}P{1.3cm}P{1.4cm}P{1.5cm}P{1.5cm}}
\hline
\hline
    \multirow{2}{*}{\textbf{Model}} & \textbf{SNLI} & \textbf{Multi-NLI} & \textbf{Quora} & \textbf{Twitter-URL} & \textbf{PIT-2015} & \textbf{STS-2014} & \textbf{WikiQA} & \textbf{TrecQA}  \\ 
    & Acc & Acc\_m/Acc\_um & Acc & F1 & F1 & \(r\) & MAP/MRR & MAP/MRR \\ 
\hline
InferSent & 0.845 & -/- & - & - &- & 0.700\footnotemark[5] &- &- \\ 
\hline
SSE & 0.860 & 0.746/0.736 & - & - &- & - &- &- \\ 
\hline
DecAtt & 0.863 & - & 0.865\footnotemark[3] & - &- & - &- &- \\ 
\hline
ESIM$_{tree}$ & 0.878 & - & - & - &- & - &- &- \\ 
\hline
ESIM$_{seq}$ & 0.880 & 0.723/0.721\footnotemark[4] & - & - &- & - &- &- \\ 
\hline
ESIM$_{seq+tree}$ & 0.886 & - & - & - &- & - &- &- \\ 
\hline
PWIM & - & -& - & 0.749 & 0.667 & 0.767 & 0.709/0.723 & 0.759/0.822 \\ 
\hline
\end{tabular}

\caption{Reported results from original papers, which are mostly limited to a few datasets. For the Multi-NLI dataset, Acc\_m represents testing accuracy for the matched genre and Acc\_um for the unmatched genre.  } 
\label{model_reported_results}
\end{table*}
\egroup

\bgroup
\def\arraystretch{1.8}
\begin{table*}[ht!]
\small
\centering
\begin{tabular}{P{1.6cm}P{1cm}P{1.6cm}P{1cm}P{1.8cm}P{1.3cm}P{1.4cm}P{1.5cm}P{1.5cm}}
\hline
\hline
    \multirow{2}{*}{\textbf{Model}} & \textbf{SNLI} & \textbf{Multi-NLI} & \textbf{Quora} & \textbf{Twitter-URL} & \textbf{PIT-2015} & \textbf{STS-2014} & \textbf{WikiQA} & \textbf{TrecQA}  \\ 
    & Acc & Acc\_m/Acc\_um & Acc & F1 & F1 & \(r\) & MAP/MRR & MAP/MRR \\ 
\hline
InferSent & 0.846 & 0.705/0.703 & \underline{0.866} & 0.746 & 0.451 & \underline{0.715} & 0.287/0.287 & 0.521/0.559 \\ 
\hline
SSE & 0.855 & 0.740/0.734 & \textbf{0.878} & 0.650 & 0.422 & 0.378 & 0.624/0.638 & 0.628/0.670 \\ 
\hline
DecAtt & 0.856 & 0.719/0.713 & 0.845 & 0.652 & 0.430 & 0.317 & 0.603/0.619 & 0.660/0.712 \\ 
\hline
ESIM$_{tree}$ & 0.864 & 0.736/0.727 & 0.755 & 0.740 & 0.447 &  0.493 & 0.618/0.633 & 0.698/0.734 \\ 
\hline
ESIM$_{seq}$  & \underline{0.870} & \underline{0.752/0.738} & 0.850 & 0.748 & 0.520 & 0.602 & \underline{0.652/0.664} & \textbf{0.771/0.795} \\ 
\hline
ESIM$_{seq+tree}$ & \textbf{0.871} & \textbf{0.753/0.748} & 0.854 & \underline{0.759} & \underline{0.538} & 0.589 & 0.647/0.658 & 0.749/0.768 \\ 
\hline
PWIM & 0.822 & 0.722/0.716 & 0.834 & \textbf{0.761} & \textbf{0.656} & \textbf{0.743} & \textbf{0.706/0.723} & \underline{0.739/0.795} \\ 
\hline
\end{tabular}
\caption{Replicated results with our reimplementation in PyTorch across multiple tasks and datasets. The best result in each dataset is denoted by a \textbf{bold} typeface, and the second best is denoted by an \underline{underline}.} 
\vspace{-.1in}
\label{model_replicated_results}
\end{table*}
\egroup

\vspace{-.1in}
\noindent\textbf{How important is attentive interaction for sentence pair modeling? Why does SSE excel on Quora?} \\
Both ESIM and DecAtt (Eq. \ref{eq:soft_alignment}) calculate an attention-based soft alignment between a sentence pair, which was also proposed in \cite{rocktaschel2015reasoning} and \cite{wang2016compare} for sentence pair modeling, whereas PWIM utilizes a hard attention mechanism. Both attention strategies are critical for model performance. In PWIM model \cite{he-lin:2016:N16-1}, we observed a 1$\sim$2 point performance drop after removing the hard attention, 0$\sim$3 point performance drop and $\sim$25\% training time reduction after removing the 19-layer CNN aggregation. Likely without even the authors of SSE knowing, the SSE model performs extraordinarily well on the Quora corpus, perhaps because Quora contains many sentence pairs with less complicated inter-sentence interactions (e.g., many identical words in the two sentences) and incorrect ground truth labels (e.g., \textit{What is your biggest regret in life?} and \textit{What's the biggest regret you've had in life?} are labeled as non-duplicate questions by mistake).\\



\vspace{-.1in}
\subsubsection{Learning Curves and Training Time}
\begin{figure}[ht!]
\begin{subfigure}{0.24\textwidth}
\includegraphics[width=\linewidth, height=3.5cm]{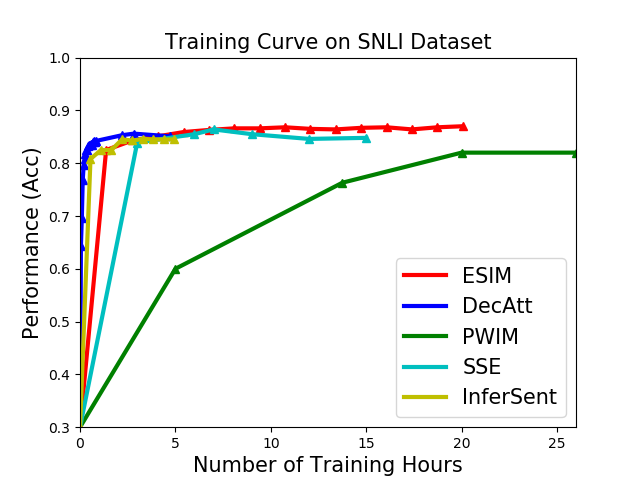} 
\end{subfigure}
\begin{subfigure}{0.24\textwidth}
\includegraphics[width=\linewidth, height=3.5cm]{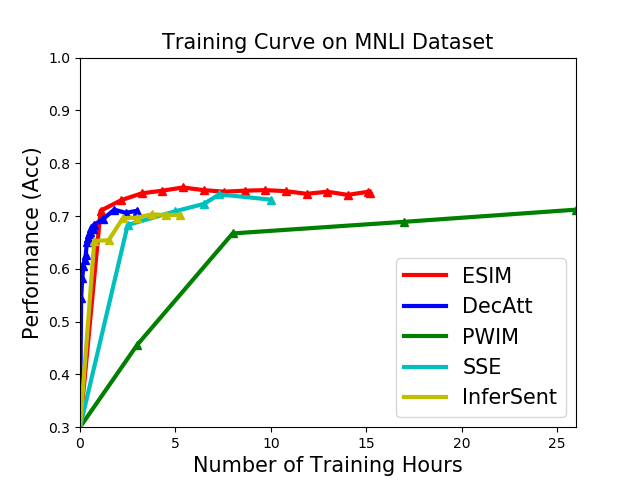}
\end{subfigure}
\begin{subfigure}{0.24\textwidth}
\includegraphics[width=\linewidth, height=3.5cm]{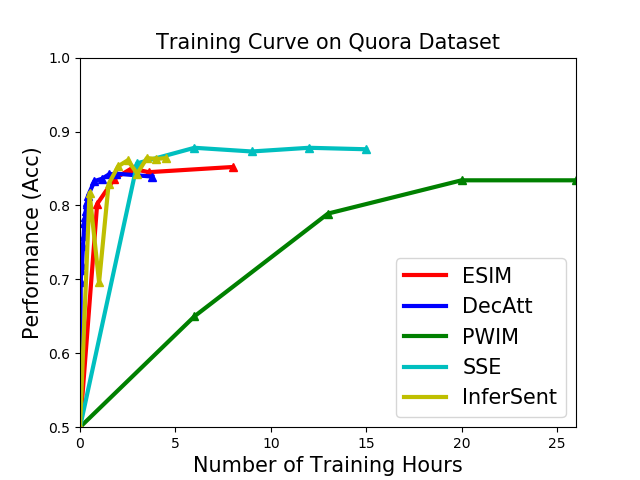}
\end{subfigure}
\begin{subfigure}{0.24\textwidth}
\includegraphics[width=\linewidth, height=3.5cm]{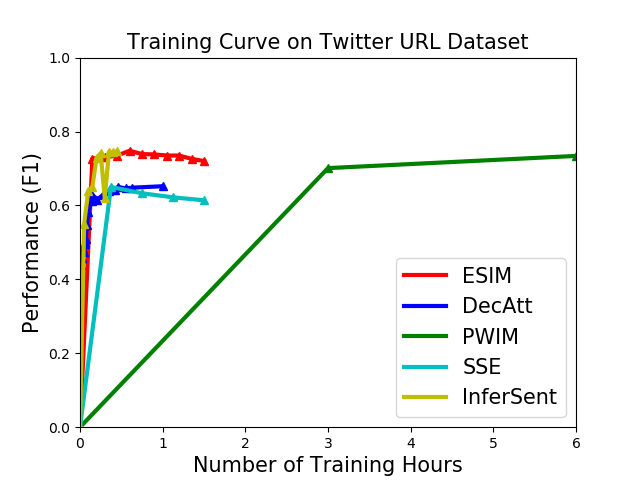}
\end{subfigure}

\begin{subfigure}{0.24\textwidth}
\includegraphics[width=\linewidth, height=3.5cm]{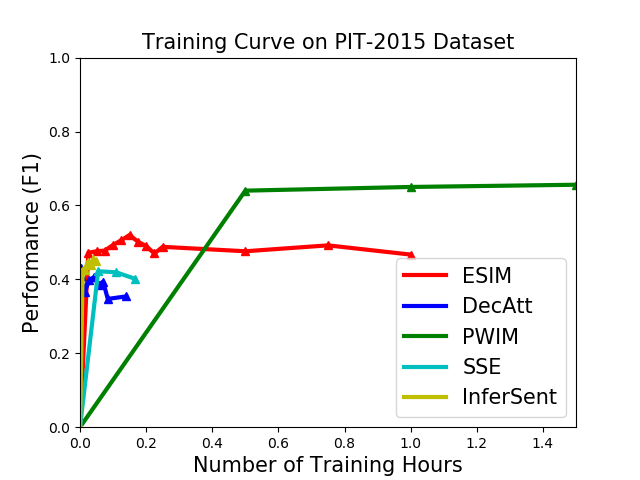}
\end{subfigure}
\begin{subfigure}{0.24\textwidth}
\includegraphics[width=\linewidth, height=3.5cm]{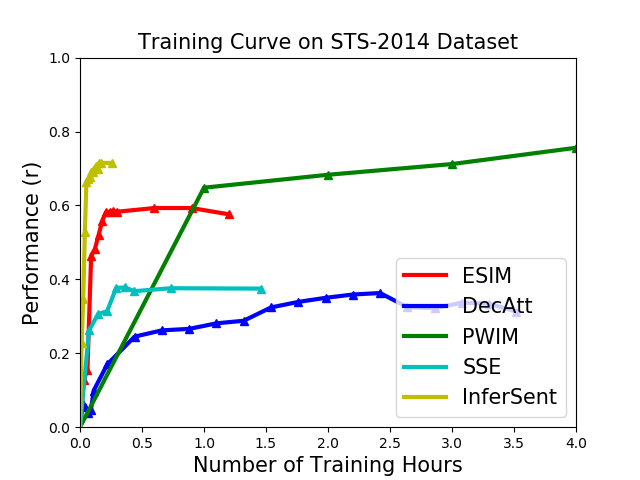}
\end{subfigure}
\begin{subfigure}{0.24\textwidth}
\includegraphics[width=\linewidth, height=3.5cm]{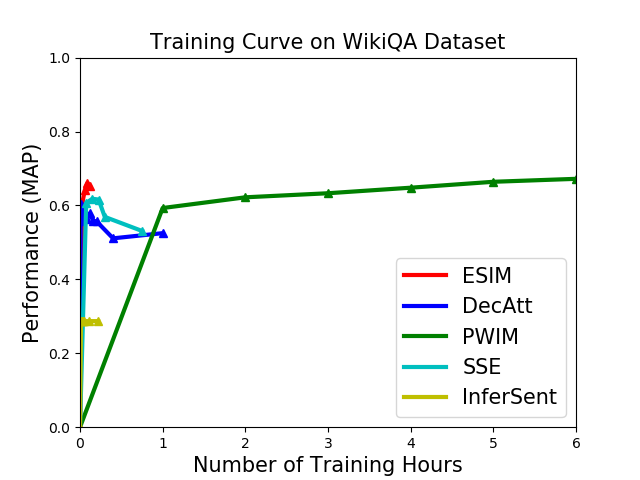}
\end{subfigure}
\begin{subfigure}{0.24\textwidth}
\includegraphics[width=\linewidth, height=3.5cm]{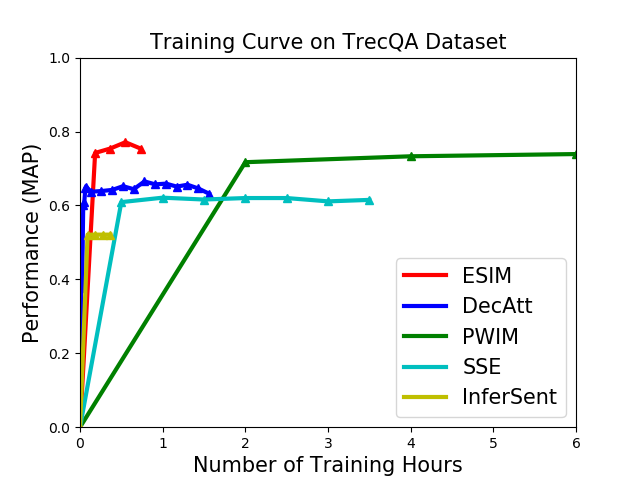}
\end{subfigure}
\caption{Training curves of ESIM, DecAtt, PWIM, SSE and InferSent models on eight datasets. }
\label{fig:Training_curve}
\end{figure}

Figure \ref{fig:Training_curve} shows the learning curves. The DecAtt model converges quickly and performs well on large NLI datasets due to its design simplicity. PWIM is the slowest model (see time comparison in Table \ref{tab:training_timel}) but shows very strong performance on semantic similarity and paraphrase identification datasets. ESIM and SSE keep a good balance between training time and performance.

\footnotetext[3]{This number was reported in \cite{tomar-EtAl:2017:SCLeM} by co-authors of DecAtt \cite{parikh-EtAl:2016:EMNLP2016}. }
\footnotetext[4]{This number was reproduced by Williams et al. \shortcite{williams2017broad}.}
\footnotetext[5]{This number was generated by InferSent traind on SNLI and Multi-NLI datasets.}

\begin{table*}[ht!]
\small
    \centering
    \begin{tabular}{P{5.5cm}P{1.25cm}P{1.25cm}P{1.25cm}P{1.25cm}P{1.25cm}P{1.25cm}}
     \hline
    \hline
     & \textbf{InferSent} & \textbf{SSE} & \textbf{DecAtt} & \textbf{ESIM$_{seq}$} & \textbf{ESIM$_{tree}$} & \textbf{PWIM} \\
     \hline
     Number of parameters & 47M & 140M & 380K & 4.3M & 7.7M & 2.2M \\
    \hline
    Avg epoch time (seconds) / sentence pair  & 0.005 & 0.032 & 0.0006 & 0.013 & 0.016 & 0.60  \\
    \hline
    Ratio compared to DecAtt model  & $\times$8 & $\times$53 & 1 & $\times$22 & $\times$26 & $\times$1000  \\
    \hline
    \end{tabular}
    \caption{Average training time per sentence pair in the Twitter-URL dataset (similar time for other datasets).}
    \label{tab:training_timel}
\end{table*}

\subsubsection{Effects of Training Data Size}
\begin{figure}
\begin{floatrow}
\ffigbox{%
  \includegraphics[width=65mm]{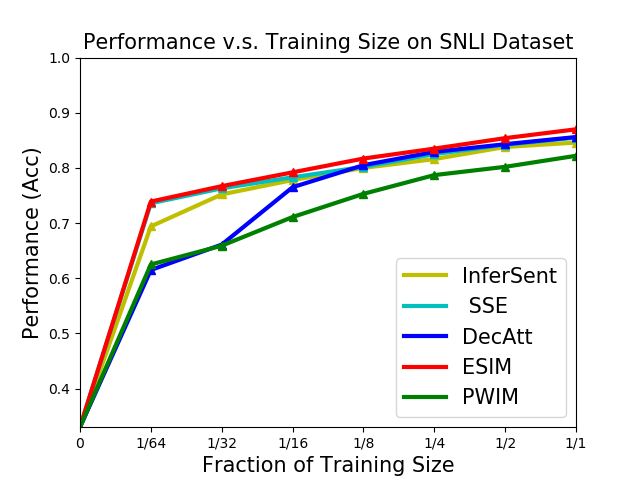}%
}{%
  \caption{Performance vs. training size (log scale in x-axis) on SNLI dataset.}%
  \label{fig:performance_vs_training_size}
}
\capbtabbox{%
\small
\begin{tabular}{P{1cm}P{1cm}P{1cm}P{1cm}|P{1.3cm}}
\hline
\hline
 \multirow{2}{*}{\textbf{Models}} & \textbf{Quora} & \textbf{URL} & \textbf{PIT} & train/test\\
 \cline{2-4}
 & \multicolumn{3}{c|}{trained on Quora} & on PIT\\
 \hline
 InferSent & 0.866 & 0.528 & 0.394 & 0.451\\
 SSE & 0.878 & 0.681 & 0.594  & 0.422\\
 DecAtt & 0.845 & 0.649 & 0.497 & 0.430\\
 ESIM$_{seq}$ &  0.850 & 0.643 & 0.501 & 0.520\\
 PWIM & 0.835 & 0.601 & 0.518 & 0.656\\
\cline{1-4}
 &  \multicolumn{3}{c|}{trained on URL} & \\
\cline{1-4}
 InferSent & 0.703 & 0.746 & 0.535 & 0.451\\
 SSE & 0.630 & 0.650 & 0.477 & 0.422 \\
 DecAtt & 0.632 & 0.652 & 0.450 & 0.430\\
 ESIM$_{seq}$ & 0.641 & 0.748 & 0.511 & 0.520\\
 PWIM & 0.678 & 0.761 & 0.634 & 0.656\\
\hline
\end{tabular}
}{%
  \caption{Transfer learning experiments for paraphrase identification task.}%
  \label{transfer_learning_results}
}
\end{floatrow}
\end{figure}

As shown in Figure \ref{fig:performance_vs_training_size}, we experimented with different training sizes of the largest  SNLI dataset. All the models show improved performance as we increase the training size. ESIM and SSE have very similar trends and clearly outperform PWIM on the SNLI dataset. DecAtt shows a performance jump when the training size exceeds a threshold.

\subsubsection{Categorical Performance Comparison}
We conducted an in-depth analysis of model performance on the Multi-domain NLI dataset based on different categories: text genre, sentence pair overlap, and sentence length. As shown in Table \ref{categorical_performance_analyses}, all models have comparable performance between matched genre and unmatched genre. Sentence length and overlap turn out to be two important factors -- the longer the sentences and the fewer tokens in common, the more challenging it is to determine their semantic relationship. These phenomena shared by the state-of-the-art systems reflect their similar design framework which is symmetric at processing both sentences in the pair, while question answering and natural language inference tasks are directional \cite{ghaeini2018dr}. How to incorporate asymmetry into model design will be worth more exploration in future research.

\begin{table*}[ht]
\small
\centering
\begin{tabular}{P{1.2cm}P{1.9cm}P{1.7cm}P{1.7cm}P{1.7cm}P{1.7cm}P{1.7cm}P{1.7cm}}
\hline
\hline
  &  \textbf{Category} & \textbf{\#Examples} & \textbf{InferSent} & \textbf{SSE} & \textbf{DecAtt} & \textbf{ESIM$_{seq}$} & \textbf{PWIM}  \\ 
  \hline
\multirow{4}{*}{Matched} & Fiction & 1973 & 0.703 & 0.727 & 0.706 & 0.742 & 0.707 \\
\multirow{4}{*}{Genre}  & Government & 1945 & 0.753 & 0.746 & 0.743 & 0.790 & 0.751 \\
  & Slate & 1955 & 0.653 & 0.670 & 0.671 & 0.697 & 0.670\\
  & Telephone & 1966 & 0.718 & 0.728 & 0.717 & 0.753 & 0.709\\
  & Travel & 1976 & 0.705 & 0.701 & 0.733 & 0.752 & 0.714\\
\hline
\multirow{4}{*}{Mismatched} & 9/11 & 1974 & 0.685 & 0.710 & 0.699 & 0.737 & 0.711\\
\multirow{4}{*}{Genre}  & Face-to-face & 1974 & 0.713 & 0.729 & 0.720 & 0.761 & 0.710\\
  & Letters & 1977 & 0.734 & 0.757 & 0.754 & 0.775 & 0.757\\
  & OUP & 1961 & 0.698 & 0.715 & 0.719 & 0.759 & 0.710\\
  & Verbatim & 1946 & 0.691 & 0.701 & 0.709 & 0.725 & 0.713\\
\hline
\multirow{3}{*}{Overlap} & $>$60\% & 488 & 0.756 & 0.795 & 0.805 & 0.842 & 0.811\\
 & 30\% $\sim$ 60\% & 3225 & 0.740 & 0.751 & 0.745 & 0.769 & 0.743\\
 & $<$30\% & 6102 & 0.685 & 0.689 & 0.691 & 0.727 & 0.682\\
\hline
\multirow{3}{*}{Length} & $>$20 tokens & 3730 & 0.692 & 0.676 & 0.685 & 0.731 & 0.694\\
 & 10$\sim$20 tokens & 3673 & 0.712 & 0.725 & 0.721 & 0.753 & 0.720\\
 & $<$10 tokens & 2412 & 0.721 & 0.758 & 0.748 & 0.762 & 0.724\\
\hline
\end{tabular}
\caption{Categorical performance (accuracy) on Multi-NLI dataset. Overlap is the percentage of shared tokens between two sentences. Length is calculated based on the number of tokens of the longer sentence.} 
\vspace{-.1in}
\label{categorical_performance_analyses}
\end{table*}

\subsubsection{Transfer Learning Experiments}
In addition to the cross-domain study (Table \ref{categorical_performance_analyses}), we conducted transfer learning experiments on three paraphrase identification datasets (Table \ref{transfer_learning_results}). The most noteworthy phenomenon is that the SSE model performs better on Twitter-URL and PIT-2015 when trained on the large out-of-domain Quora data than the small in-domain training data. Two likely reasons are: (i) the SSE model with over 29 million parameters is data hungry and (ii) SSE model is a sentence encoding model, which generalizes better across domains/tasks than sentence pair interaction models. Sentence pair interaction models may encounter difficulties on Quora, which contains sentence pairs with the highest word overlap (51.5\%) among all datasets and often causes the interaction patterns to focus on a few key words that differ. In contrast, the Twitter-URL dataset has the lowest overlap (23.0\%) with a semantic relationship that is mainly based on the intention of the tweets.


\section{Conclusion}
We analyzed five different neural models (and their variations) for sentence pair modeling and conducted a series of experiments with eight representative datasets for different NLP tasks. We quantified the importance of the LSTM encoder and attentive alignment for inter-sentence interaction, as well as the transfer learning ability of sentence encoding based models. We showed that the SNLI corpus of over 550k sentence pairs cannot saturate the learning curve. We systematically compared the strengths and weaknesses of different network designs and provided insights for future work.

\section*{Acknowledgements}

We thank Ohio Supercomputer Center \cite{Oakley2012} for computing resources. This work was supported in part by NSF CRII award (RI-1755898) and DARPA through the ARO (W911NF-17-C-0095). The content of the information in this document does not necessarily reflect the position or the policy of the U.S. Government, and no official endorsement should be inferred. 


\bibliographystyle{coling2018}
\bibliography{coling2018}

\begin{appendices}
\section{Pretrained Word Embeddings}
We used the 200-dimensional GloVe word vectors \cite{pennington2014glove}, trained on 27 billion words from Twitter (vocabulary size of 1.2 milion words) for Twitter URL \cite{lan2017continuously} and PIT-2015 \cite{xu2015semeval} datasets, and the 300-dimensional GloVe vectors, trained on 840 billion words (vocabulary size of 2.2 milion words) from Common Crawl for all other datasets. For out-of-vocabulary words, we initialized the word vectors using normal distribution with mean \(0\) and deviation \(1\). 

\section{Hyper-parameter Settings}
We followed original papers or code implementations to set hyper-parameters for these models. In Infersent model \cite{conneau-EtAl:2017:EMNLP2017}, the hidden dimension size for Bi-LSTM is 2048, and the fully connected layers have 512 hidden units. In SSE model \cite{nie-bansal:2017:RepEval}, the hidden size for three Bi-LSTMs is 512, 2014 and 2048, respectively. The fully connected layers have 1600 units. PWIM \cite{he-lin:2016:N16-1} and ESIM \cite{Chen-Qian:2017:ACL} both use Bi-LSTM for context encoding, having 200 hidden units and 300 hidden units respectively. The DecAtt model \cite{parikh-EtAl:2016:EMNLP2016} uses three kinds of feed forward networks, all of which have 300 hidden units. Other parameters like learning rate, batch size, dropout rate, and all of them use the same settings as in original papers.

\section{Fine-tuning the Models}
It is not practical to fine tune every hyper-parameter in every model and every dataset, since we want to show how these models can generalize well on other datasets, we need try to avoid fine-tuning these parameters on some specific datasets, otherwise we can easily get over-fitted models. Therefore, we keep the hyper-parameters unchanged across different datasets, to demonstrate the generalization capability of each model. The default number of epochs for training these models is set to 20, if some models could converge earlier (no more performance gain on development set), we would stop running them before they approached epoch 20. The 20 epochs can guarantee every model get converged on every dataset. 

\end{appendices}

\end{document}